\documentclass[letterpaper, 10 pt, conference]{ieeeconf}  

\IEEEoverridecommandlockouts                              

\usepackage{graphics} 
\usepackage{epsfig} 
\usepackage{times} 
\usepackage{amsmath} 
\usepackage{amssymb}  
\usepackage{hyperref}
\usepackage{array}
\usepackage{url}
\usepackage{subfigure}
\usepackage{booktabs}
\usepackage{caption}

\title{\LARGE \bf
Optimizing the 
extended Fourier Mellin Transformation Algorithm
}

\author{Wenqing Jiang$^{1*}$, Chengqian Li$^{1*}$, Jinyue Cao$^{1}$ and S\"{o}ren Schwertfeger$^{1}$
\thanks{$^{*}$ indicates equal contribution.}%
\thanks{$^{1}$Wenqing Jiang, Chengqian Li, Jinyue Cao, and S\"{o}ren Schwertfeger are with the School of Information Science and Technology, ShanghaiTech University, Shanghai, China
        {\tt\small \{jiangwq, lichq, caojy, soerensch\}@shanghaitech.edu.cn}}%
}

\begin{document}

\maketitle
\thispagestyle{empty}
\pagestyle{empty}

\begin{abstract}

    With the increasing application of robots, stable and efficient Visual Odometry (VO) algorithms are becoming more and more important. Based on the Fourier Mellin Transformation (FMT) algorithm, the extended Fourier Mellin Transformation (eFMT) is an image registration approach that can be applied to downward-looking cameras, for example on aerial and underwater vehicles. eFMT extends FMT to multi-depth scenes and thus more application scenarios. It is a visual odometry method which estimates the pose transformation between three overlapping images. On this basis, we develop an optimized eFMT algorithm that improves certain aspects of the method and combines it with back-end optimization for the small loop of three consecutive frames. For this we investigate the extraction of uncertainty information from the eFMT registration, the related objective function and the graph-based optimization. Finally, we design a series of experiments to investigate the properties of this approach and compare it with other VO and SLAM (Simultaneous Localization and Mapping) algorithms. The results show the superior accuracy and speed of our o-eFMT approach, which is published as open source.

\end{abstract}

\section{Introduction}

The Fourier-Mellin-Transform (FMT) algorithm, first introduced in the 1990s, is a traditional image registration algorithm for images captured with pinhole cameras. It is a popular algorithm in many fields of studies such as remote sensing~\cite{xie2019novel}, robotics~\cite{pfingsthorn2013large} and image analysis~\cite{turski2000projective, 1562722}, to name a few. The classic Fourier-Mellin transform was first presented by Reddy and Chatterji~\cite{reddy1996fft}, and over the past few decades, improved massively on its computational efficiency and robustness~\cite{bulow2009fast,bulow2009online, DERRODE200157}. A detailed review of Fourier-based image correlation is provided in \cite{8844710}.

FMT is based on Fourier transform analysis and uses a phase-only matched filter~\cite{chen1994symmetric} to estimate the rotation and translation between two images. This is in contrast to currently more popular VO algorithms, which often use either feature extraction and matching for image registration such as ORB-SLAM3~\cite{ORBSLAM3_2020}, or rely on direct methods such as DSO~\cite{Engel-et-al-pami2018}. The mainstream methods work just fine until feature-deprived or highly repetitive environments occur, and that is where FMT shows superior performance~\cite{xu2019improved, 387491, 4290156, 4584311, 10.1007/978-3-642-13408-1_14}. Yet, this algorithm has its weak spot. One specific aspect of FMT is, that it can only estimate camera motions with 4 Degrees of Freedom (DOF): given that the image lies in the $XY$ plane, the camera can translate in the $x$-, $y$-, $z$-axis and rotate around the $z$-axis, but rotation around the $x$- or $y$-axis (roll and pitch, respectively) are not allowed. This limitation restricts the application scenarios of FMT to systems that utilize down-looking cameras without roll or pitch. Examples for those are down-looking cameras on satellites \cite{le2011image}, aerial vehicles \cite{grabe2015nonlinear,sa2018weedmap} or underwater vehicles \cite{pfingsthorn2013large}, which are still exciting areas for FMT to shine. Additionally, FMT can be used as part of a more complex VO system, e.g. with omni-directional cameras \cite{kuang2019pose,xu2019improved, 4587553}.

Another major shortcoming of FMT is the fact that it requires the depicted scene to be flat and parallel to the camera, which means that the environment needs to be planar and parallel to the imaging plane. In other words, only single depth is allowed. Our recent work~\cite{xu2021rethinking} eFMT overcomes this problem and is able to remove the constraints of equidistance and planar environment. eFMT, short for extended Fourier-Mellin-Transform, extends the algorithm application to general multi-depth environments. This is major breakthrough, as for the first time, it allows this spectral-based method to be applied to any environment, no strings attached. eFMT points out that if the depths of objects are different, the pixels' motion will also be different even if the camera's motion is the same. This is due to perspective projection. Since FMT can only estimate the image motion in the dominant depth, the camera's speed could not be correctly inferred when the dominant depth changes. eFMT first represents the translation in a one-dimensional translation energy vector obtained from the phase shift diagram instead of just picking the maximum peak, as does the classical FMT. It then puts the zoom and translation in the same reference frame based on pattern matching, and finally, assigns a magnitude (change of camera speed) to the second of the two found unit translation vectors of three consecutive frames.
Their work shows the excellent performance of eFMT in comparison to FMT and also traditional methods like ORB-SLAM3, SVO \cite{forster2016svo} and DSO.

In this paper, we propose an optimized eFMT (o-eFMT), which modifies the eFMT method and adds a back-end optimization to it. First of all, our method shows superior performance to theirs. As for back-end optimization, eFMT is still a VO algorithm, and like all VO algorithms, it only considers the correlation between adjacent timestamps, while errors can be accumulated over time and lead to unreliable results. Our method, o-eFMT, like eFMT, considers three consecutive frames, but not only the transformations between the first two frames and the last two frames, we also consider the transformation between the first and last frame and add a constraint to these three transformations, additionally estimate their uncertainty and use all this to optimize the original results. 


\begin{figure*}[t]
    \centering
    \includegraphics[width=0.95\linewidth]{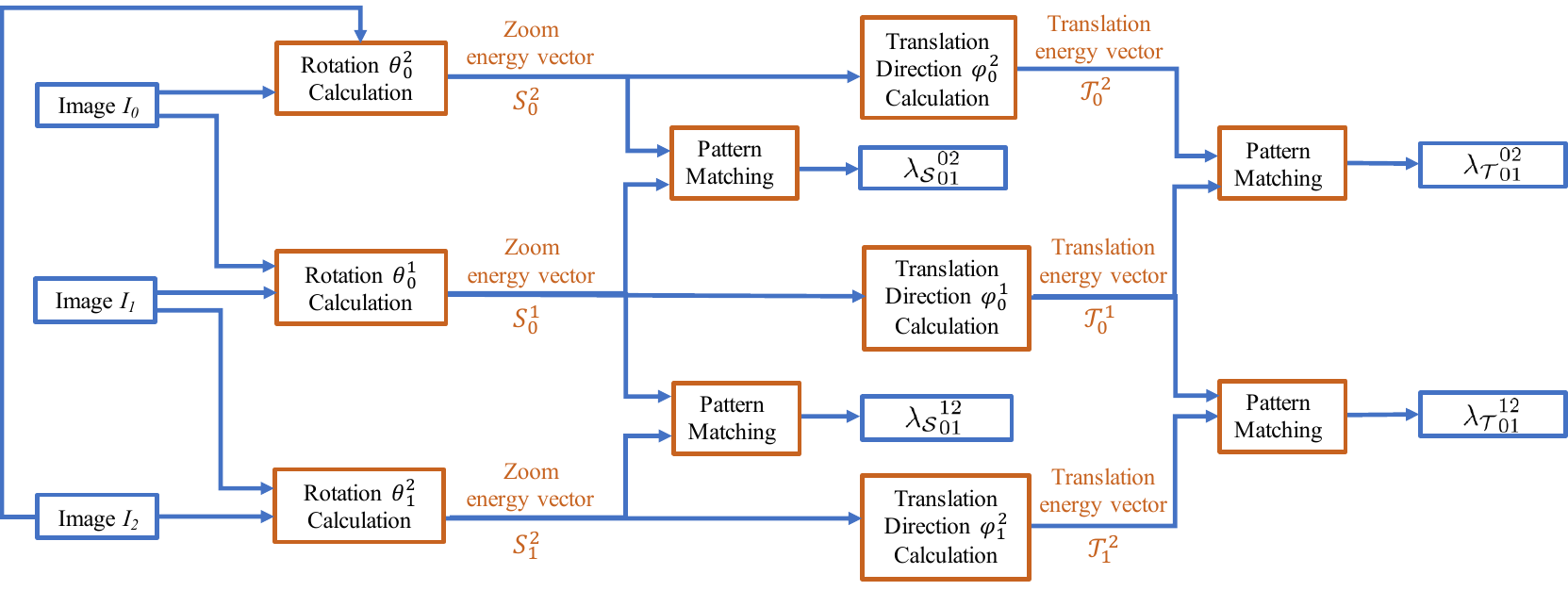}
    \caption{The pipeline of o-eFMT. The orange blocks are where we modified and 
 complemented eFMT.}
    \label{fig:pipeline}
\end{figure*}

The contributions of this paper are summarized as follows:
\begin{enumerate}
    \item In phase shift diagram processing, we extract the zoom energy vector in a simpler way to bypass the multi-zoom calculation. This drastically reduces the computational complexity of multiple Fourier transforms in the presence of motion on the $z$-axis in a multi-depth environment. We further unify the extraction method of both translation energy vector and zoom energy vector.
    \item In pattern matching, we reduce the computation cost by using tighter bounds of possible scalings and perform Gaussian filtering on the energy vectors and Laplace transform on the error sequence of all probable factors to reduce the influence of noise and for improved matching accuracy. 
    \item We add uncertainty estimation of the rotation and translation directions for the fusion step of energy vector extraction and additional uncertainty estimations for the factors estimated by pattern matching. 
    \item For robustness enhancements, we add a local optimizer to the VO structure, which further improves the accuracy and stability.
    \item We provide the source-code of o-eFMT as well as the new datasets\footnote{\url{https://github.com/STAR-Center/o-eFMT}}.
\end{enumerate}

\section{Related work}
Vision-based positioning methods are usually divided into three categories according to how they describe the environment: feature-based, appearance-based and hybrid methods~\cite{scaramuzza2011visual}. Feature-based VO needs to extract different descriptive regions from the image and establish corresponding descriptors \cite{Mur_Artal_2015,Rosten_2010,790410} and has various other applications, e.g. \cite{chavez2019adaptive}. The appearance-based methods do not need to extract features. They depend on whole or part of the image. The hybrid methods consider the characteristics of pixel consistency and pose estimation.

Among them, FMT~\cite{chen1994symmetric} is a traditional analysis algorithm for pin-hole camera images, originally proposed by Qin-Sheng Chen et al. in the 1990s. Based on the Fourier Transform and the phase correlation method, the FMT algorithm can be used to estimate the translation and rotation transform between images. Thus, it can be used as an alternative for VO algorithm. 
In 2021 Xu and Schwertfeger developed the extended FMT VO algorithm (eFMT) ~\cite{xu2021rethinking}, which is the basis of this paper. 
From the perspective of relative pose calculation, popular VO/VSLAM frameworks are divided into filtering-based~\cite{davison2007monoslam}, key-frame-based~\cite{ORBSLAM3_2020} and direct methods~\cite{engel2014lsd}. Another positioning method is called the semi-direct method, such as SVO~\cite{forster2016svo}. It uses the direct method in image registration, but keeps the reprojection error to a minimum in pose estimation and beam adjustment. Existing feature-based methods cannot perform feature matching correctly in some challenging scenes, such as environments with fewer features, blurred motion or underwater turbid scenes. Although direct methods perform better than feature-based methods in feature deprived scenarios, they do not work well when there are fewer textures in the environment. FMT and eFMT have been shown to have a considerably better performance in such environments~\cite{xu2019improved}.

\section{Overview of eFMT}
The \textbf{extended Fourier-Mellin Transformation (eFMT)}, is an alternative visual odometry (VO) approach aiming at extending FMT to multi-depth environments while maintaining the advantages of FMT in feature-deprived scenarios. Its pipeline is similar to that of FMT \cite{reddy1996fft} with smarter ways of processing the \textbf{\textit{phase shift diagram (PSD)}}. eFMT deals with three consecutive frames of images to extract camera motion. Image registration is firstly done on the first two frames $I_0,I_1$ and the last two frames $I_1,I_2$ to obtain the 4DOF pose: zoom, rotation, translation. Then through pattern matching, the scale consistency is maintained between both poses. In detail, given two input images $I_0$ and $I_1$, eFMT first calculates the rotation and zoom. After converting the images onto the frequency domain and using Fourier transform to obtain their spectra, applying an inverse Fourier transformation on the cross power spectrum of the spectra (phase correlation method) presents the rotation and zoom PSD based on $I_0$ and $I_1$. eFMT extracts the \textbf{\textit{zoom energy vector ($\mathbb{V}_z$)}} to obtain rotation and multi-zoom. Then, it uses the rotation and each zoom to re-rotate and re-zoom $I_0$ to $I'_0$, to then perform the phase correlation method on $I'_0$ and $I_1$, the \textbf{\textit{translation energy vector ($\mathbb{V}_t$)}} is extracted from the translation PSD based on $I'_0$ and $I_1$. The magnitude of this first translation estimate is 1, the unit translation, because this monocular VO approach is anyways up to an unknown scale factor. All translation energy vectors corresponding to each zoom are fused into one. Two registrations give two zoom energy vectors $\mathbb{V}_z^{01}, \mathbb{V}_z^{12}$ and two translation energy vectors $\mathbb{V}_t^{01}, \mathbb{V}_t^{12}$. Pattern matching is performed on both $\mathbb{V}_z$ and on both $\mathbb{V}_t$ respectively, a scale consistency factor and a translation consistency factor are acquired. Those scale the translation relative to the speed of the first (unit) translation.  

\section{o-eFMT}
In this section, we explain in detail how and why we modify the eFMT\cite{xu2021rethinking} algorithm. We first made changes regarding the energy vector extraction method and regarding filtering less possible cases both before and after pattern matching. Furthermore we improve the pattern matching by applying a Laplace transform. Additionally, we perform uncertainty estimation for each step. All information along with the uncertainty estimation result is put into a back-end optimizer for local optimization of the VO structure which eventually present to us a more accurate and robust result.

In section~\ref{sec:EV}, we first explain how energy extraction is performed in eFMT, and then present our Gaussian modeled zoom energy vector extraction method that allows us to focus on one dominant zoom instead of all zooms which lowers the computational complexity. In section~\ref{sec:PM}, we explain how we embed the filtering method before the original pattern matching and the Laplace transform that smoothes the noisy data afterwards. In section~\ref{sec:Opti} we present how the uncertainty estimation of the last two section is combined into a back-end optimization of this entire structure. To clarify matters, we keep the terminology consistent with that of the original eFMT paper~\cite{xu2021rethinking}.

\begin{figure}[t]
    \centering
    \includegraphics[width=0.99\linewidth]{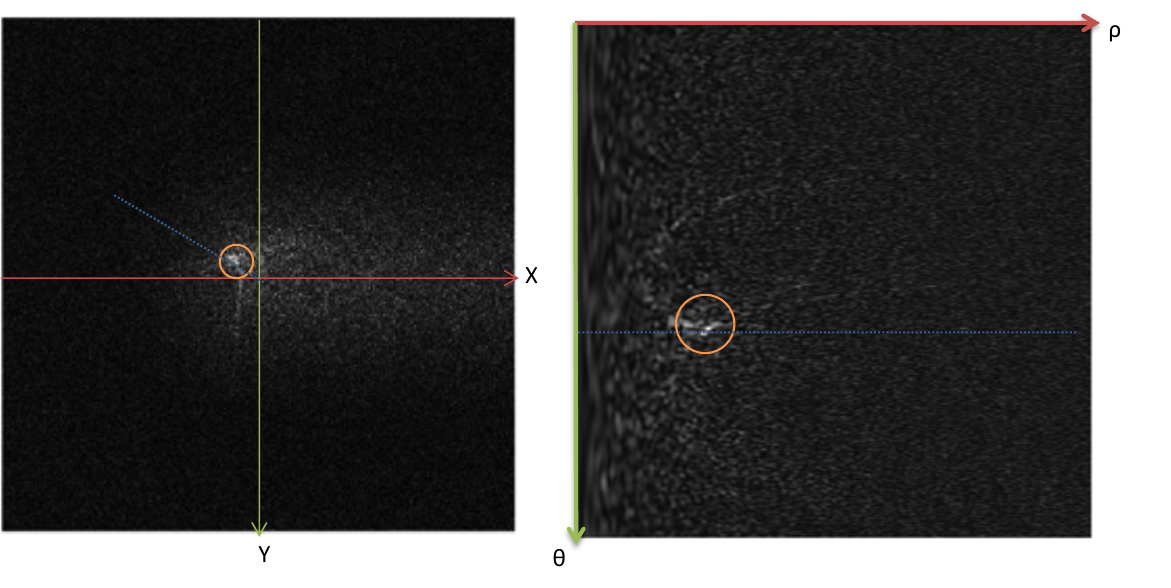}
    \caption{Transformation of the translation PSD from the polar coordinates to the Cartesian coordinates.}
    \label{fig:energy_vector}
\end{figure}

\subsection{Energy Vector}\label{sec:EV}


When recovering the rotation and zoom transformations, pixels in different depths reflect the same rotation yet with different zooms. As such, in multi-depth scenarios, the rot-scale PSD presents a row of high values. eFMT first locates this row with maximum sum energy and uniformly samples a set of multi-zoom values between the maximum zoom and the minimum zoom estimated from this row. Then, for each zoom, eFMT re-rotates and re-zooms one of the images and generates one corresponding translation PSD to recover translation information. This is repeated for each zoom factor, potentially many times. This is computationally expensive. We propose a method to bypass this multiple generation.

Due to noise and imperfect intrinsic camera calibration, the high values in the translation PSD may not locate accurately in one strict line, they could be distributed on several neighbouring rows.  Therefore, instead of sampling from the row $k$ with maximum sum energy, we set $2r$ neighbouring rows of $k$ as a block.

To be more specific, we assume that sum energy of row $[k-r, k+r]$ in the block satisfy Gaussian distribution $G(\mu, \sigma)$. Here $r=2$ is set as default. The average $\mu$ is obviously closer to the correct row corresponding to the rotation $\theta$, and $\sigma$ gives an uncertainty probability for each row in the block. Therefore, the \textit{zoom energy vector} $\mathcal{S}$ will be weighted fused according to the probability given by the Gaussian distribution as in equation \ref{eq:fusion_vector}:
\begin{equation}
    \mathcal{S} = \frac{1}{2r + 1}\sum\limits_{i=k-r}^{k+r}{\frac{1}{e^{-\frac{1}{2}{(\frac{i - \mu}{\sigma})}^2 }}}\cdot \text{row}(i) ,
    \label{eq:fusion_vector}
\end{equation}
where $\text{row}(i)$ means the $i$-th row in the \text{PSD}. This newly formed fused row is what we define as the zoom energy vector $\mathcal{S}$ and we can extract one dominant zoom from it for the next step.

While recovering the translation transformation, pixels in different depths reflect the same moving direction, yet with different lengths in movement. As such, in multi-depth scenarios, the translation PSD presents a ray of high energy starting from the center. For each zoom sampled earlier, eFMT searches for a sector in each corresponding PSD that sums up the most energy. Then eFMT samples a translation vector from this sector. The sector direction stands for the translation direction. All these translation energy vectors are finally fused into one according to the weight of the zoom energy.


We, however, process the translation PSD differently. As the high energies are located in a ray shooting from the center, we model this PSD as a polar coordinate system and translate the PSD into Cartesian coordinates as shown in Fig.~\ref{fig:energy_vector}. This leads to a new translation PSD $\mathcal{T}$ in the same format as the rot-scale PSD. We apply the same Gaussian approximation method to extract the translation energy vector as we do for the zoom energy vector. This way, we unify the energy vector extraction method. During this entire process, we only extract one $\mathcal{S}$ and one $\mathcal{T}$, while eFMT does this process multiple times, in order to reduce the computation time significantly.

This simplification leads to inaccurate motion length estimates in the rare cases where there is both: there is zooming present (camera is moving up or down) and the dominant plane changes. But our experiments show that our improved pattern matching and the optimization step more than make up for this loss in accuracy. In our future work we will address this problem in a more systematic way.

\begin{figure*}[t]
    \centering
    \subfigure[Image 0]{
        \includegraphics[width=0.22\linewidth]{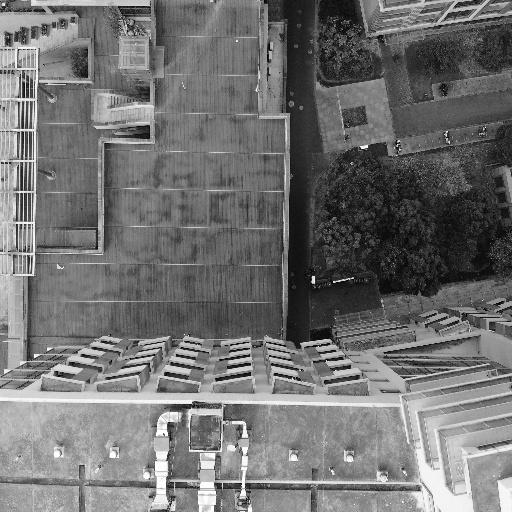}
        \label{fig:img0}
    }
    \subfigure[Image 1]{
        \includegraphics[width=0.22\linewidth]{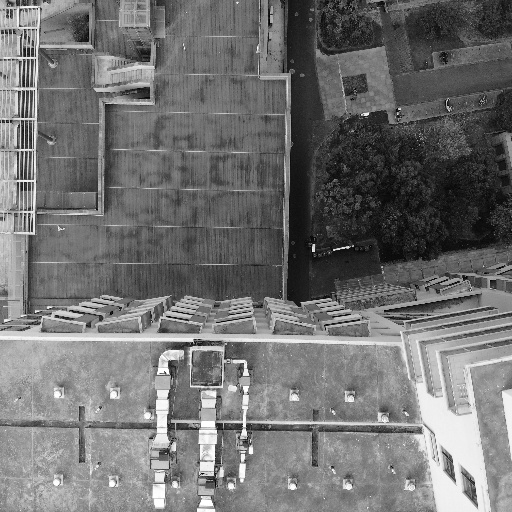}
        \label{fig:img1}
    }    
    \subfigure[Image 2]{
        \includegraphics[width=0.22\linewidth]{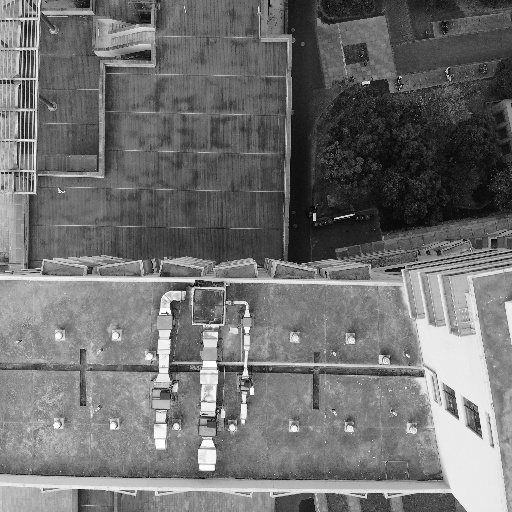}
        \label{fig:img2}
    }

    \subfigure[$\mathcal{T}_0^1$]{
        \includegraphics[width=0.22\linewidth]{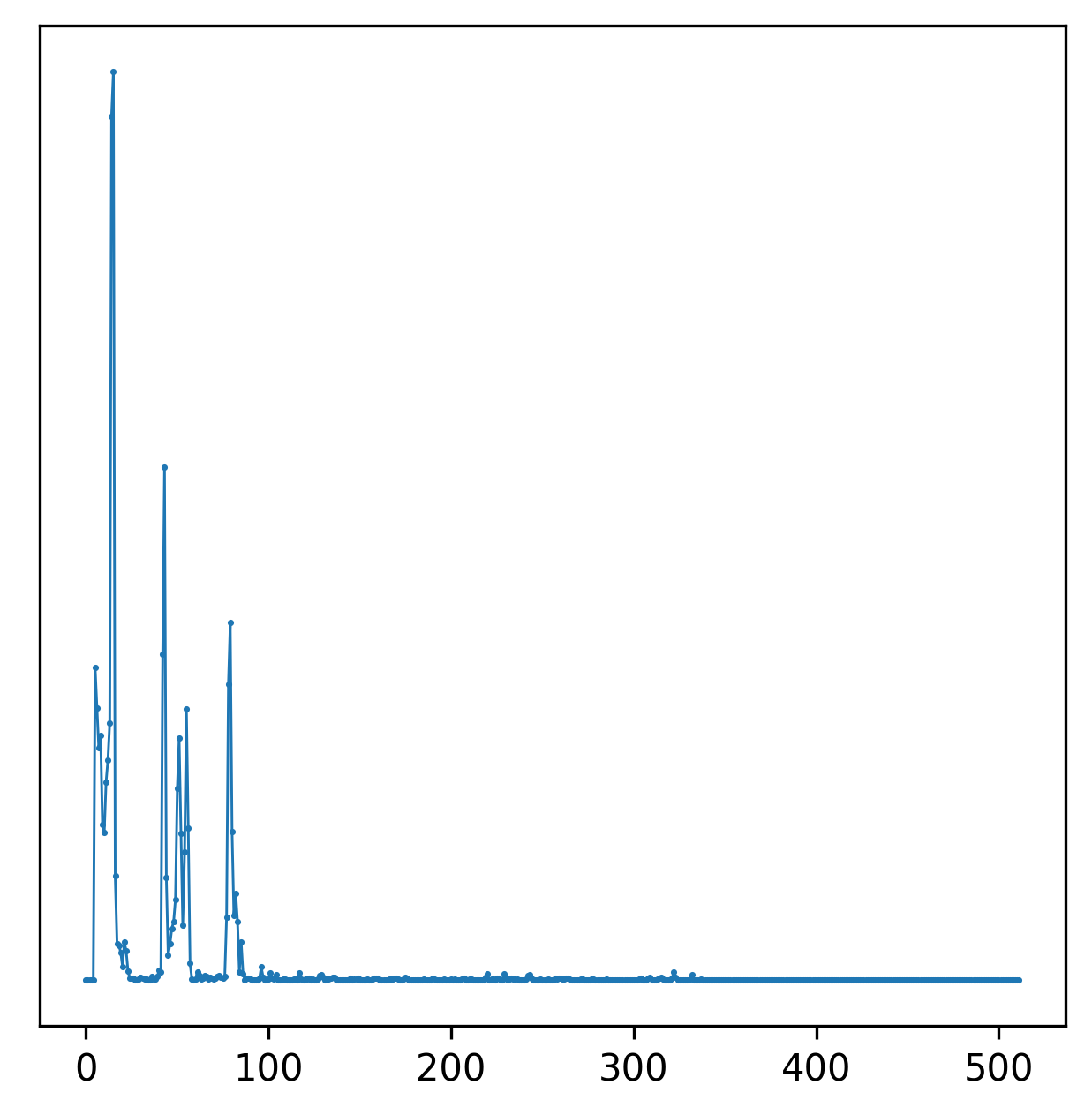}
        \label{fig:transEV01}
    }
    \subfigure[$\mathcal{T}_0^2$]{
        \includegraphics[width=0.22\linewidth]{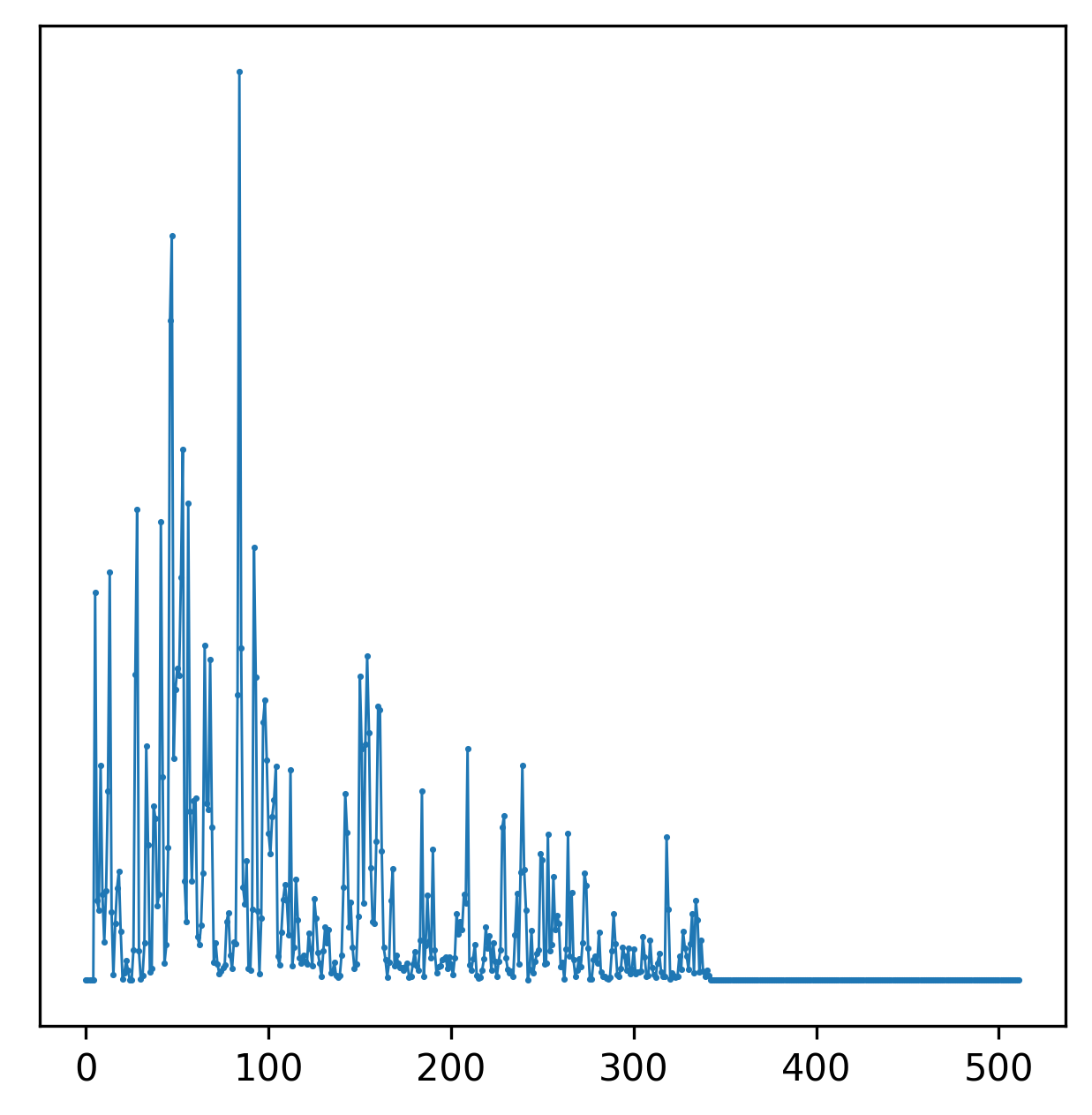}
        \label{fig:transEV02}
    }    
    \subfigure[${\lambda_{\mathcal{T}}}_{01}^{02}$]{
        \includegraphics[width=0.22\linewidth]{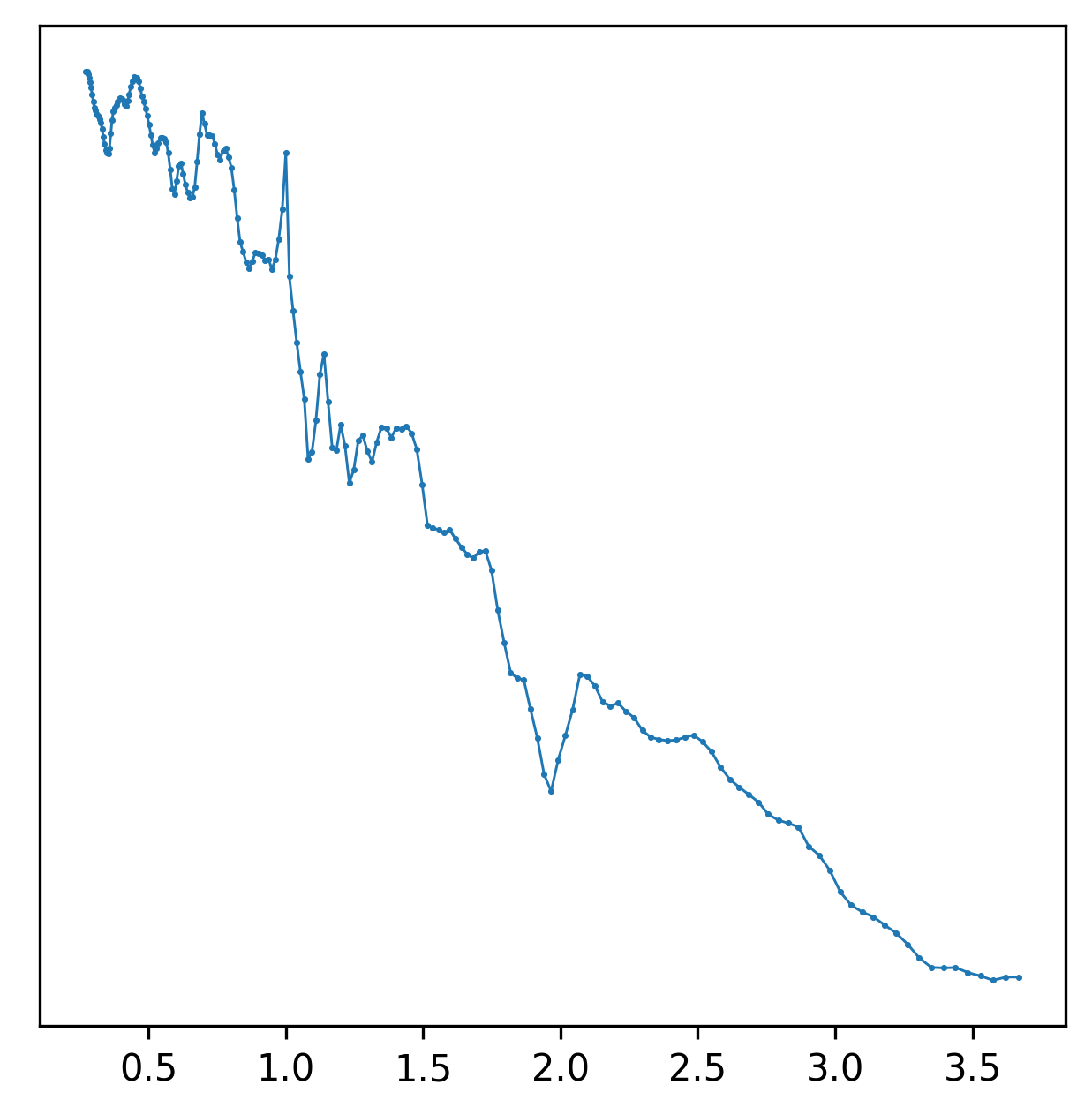}
        \label{fig:PM01_12}
    }
    \caption{Three input images (constant speed), the translation energy vectors $\mathcal{T}_0^1$ and $\mathcal{T}_0^2$ as well as the pattern matching between those. 
    The three peaks from $\mathcal{T}_0^1$ ($\triangleq$ three prominent depths in the images) can be found found at double their x-value in $\mathcal{T}_0^2$. This is a particularly difficult example. The pattern matching shows an minimum at the correct factor 2, which will be detected via the Laplace transform.}
    \label{fig:PatternMatching}
\end{figure*}

\subsection{Pattern matching}\label{sec:PM}

For multiple consecutive frames of images, the pixel depth distribution varies over time and the energy vectors change accordingly. But assuming a small camera motion, the depth distribution can be seen as fixed.
At three consecutive frames, we assume that the distribution of pixel depth is constant during two registrations. Therefore, both energy vectors should have similar structures, with only difference in scale or translation (Fig.~\ref{fig:transEV01} and~\ref{fig:transEV02}). For this reason, we use pattern matching to determine the scaling and shifting factor between energy vectors, to guarantee the scale consistency of the camera motion recovery from the image transformation process. Fig. \ref{fig:PM01_12} shows the pattern matching errors for the different scales.


For the error sequence that corresponds to the factor sequence after pattern matching, there may to be multiple minimums (see Fig.~\ref{fig:PM01_12}). But the factors that give the minimal errors are not always the correct ones. Through observation we notice that around the true factor, the errors generated by the factors in its neighbourhood change rapidly. For this reason, we apply a Laplace transform over this sequence. We find the true error that is not only minimal but also has the highest first-order derivative. Notice that we already applied a Gaussian filter over the energy vectors during their creation to smooth the data, thus the Laplace transform is not disturbed by noise too much. 

In three consecutive frames $i-2, i-1, i$, we do two registrations and for each time we get two energy vectors: the \textbf{\textit{translation energy vector}} $\mathcal{T}$ and the \textbf{\textit{scale energy vector}} $\mathcal{S}$. Define $*_{i-1}^{i}$ as the energy vector from frame $i-1$ to frame $i$, pattern matching gives us these following matching indexes:
\begin{equation}
    \begin{aligned}
        {\lambda_{\mathcal{T}}}_{i-2,i-1}^{i-1,i} &= \text{PMT}(\mathcal{T}_{i-2}^{i-1}, \mathcal{T}_{i-1}^{i})\\
        {\lambda_{\mathcal{S}}}_{i-2,i-1}^{i-1,i} &= \text{PMS}(\mathcal{S}_{i-2}^{i-1}, \mathcal{S}_{i-1}^{i})\\
    \end{aligned}
\end{equation}

PMT refers to "Pattern Matching Translation" and PMS stands for "Pattern Matching Scaling".

With these matching indexes, we can update the scale factor $s$ and translation length $\rho$ of the $i$-th registration:
\begin{equation}
    \begin{aligned}
        \rho_{i-1}^{i} &= {\lambda_{\mathcal{T}}}_{i-2,i-1}^{i-1,i} \cdot \rho_{i-2}^{i-1} \\
        s_{i-1}^{i} &= \frac{s_{i-2}^{i-1}}{ \epsilon^{{\lambda_{\mathcal{S}}}_{i-2,i-1}^{i-1,i}}  } \\
    \end{aligned}
    \label{eq:update_after_PM}
\end{equation}
where $\epsilon$ is a parameter related to the conversation of log-polar coordinate system.

Additionally, both scale energy vector and translation energy vector are related to depth, therefore, 
applying pattern matching on these two vectors can further relate the translation length on the $z$-axis to the translation length on the $XY$ plane.
We notice that only the parts that overlapped have an accurate contribution to the registration accuracy, the non-overlapping parts, however, form noise. For this reason, we add multiple strategies to boost the robustness of our algorithm.
Based on the above steps, our improved o-eFMT algorithm provides valid results in most circumstances. For further improvement in the accuracy of pose estimation, we propose to add a back-end optimization module. 
\begin{figure}[t]
    \centering
    \includegraphics[width=0.9\linewidth]{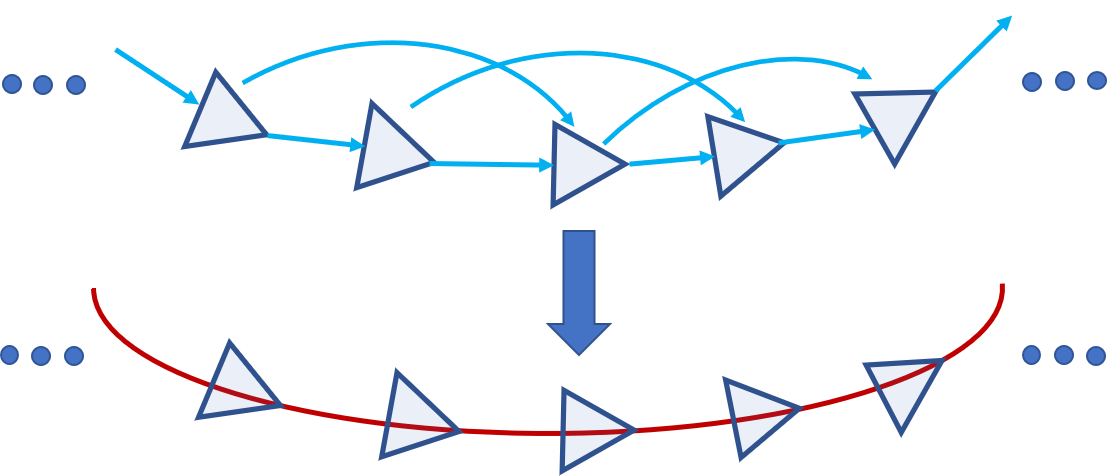}
    \caption{Visualization of the local optimization.}
    \label{fig:pose}
\end{figure}


\begin{figure*}[t]
    \centering
   \includegraphics[width=1\linewidth]{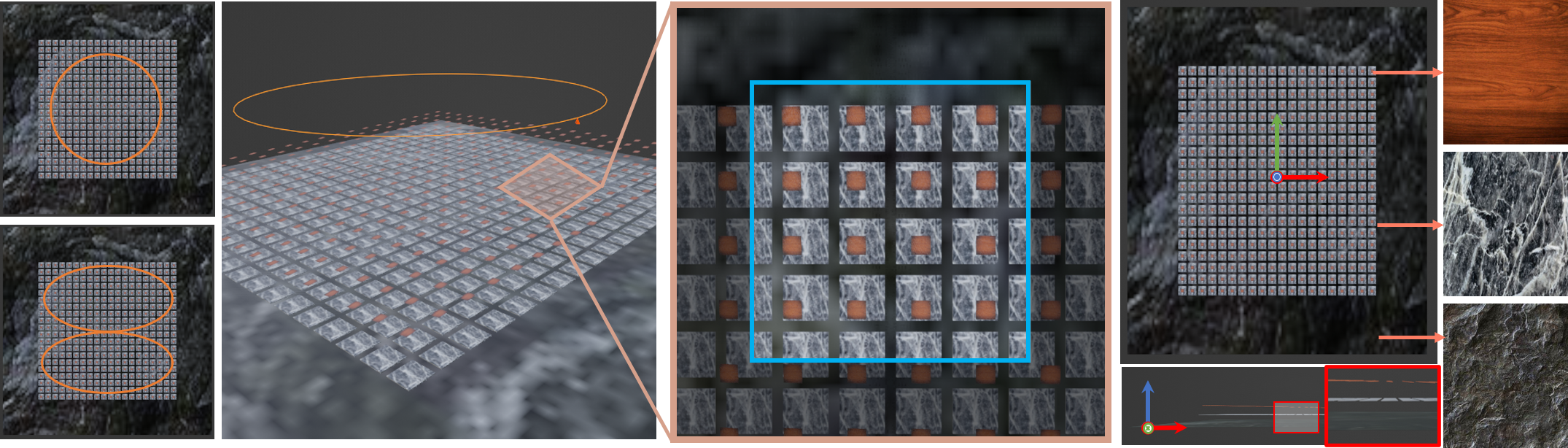}
    \caption{The scenario follows the right-hand coordinate system. It has three depth planes and each plane is made of multiple blocks. Each plane has a unique texture, shown on the right. There are two simulated camera tracks above the scenario, shown on the left. The camera follows the chosen track to generate the dataset. The yellow box shows the imaging plane obtained from the camera's perspective. According to the set camera intrinsics and the resolution of the output image, shown as the blue box, we obtain the desired dataset.}
    \label{fig:scenario}
\end{figure*}

\begin{figure}[t]
    \centering
    \includegraphics[width=1\linewidth]{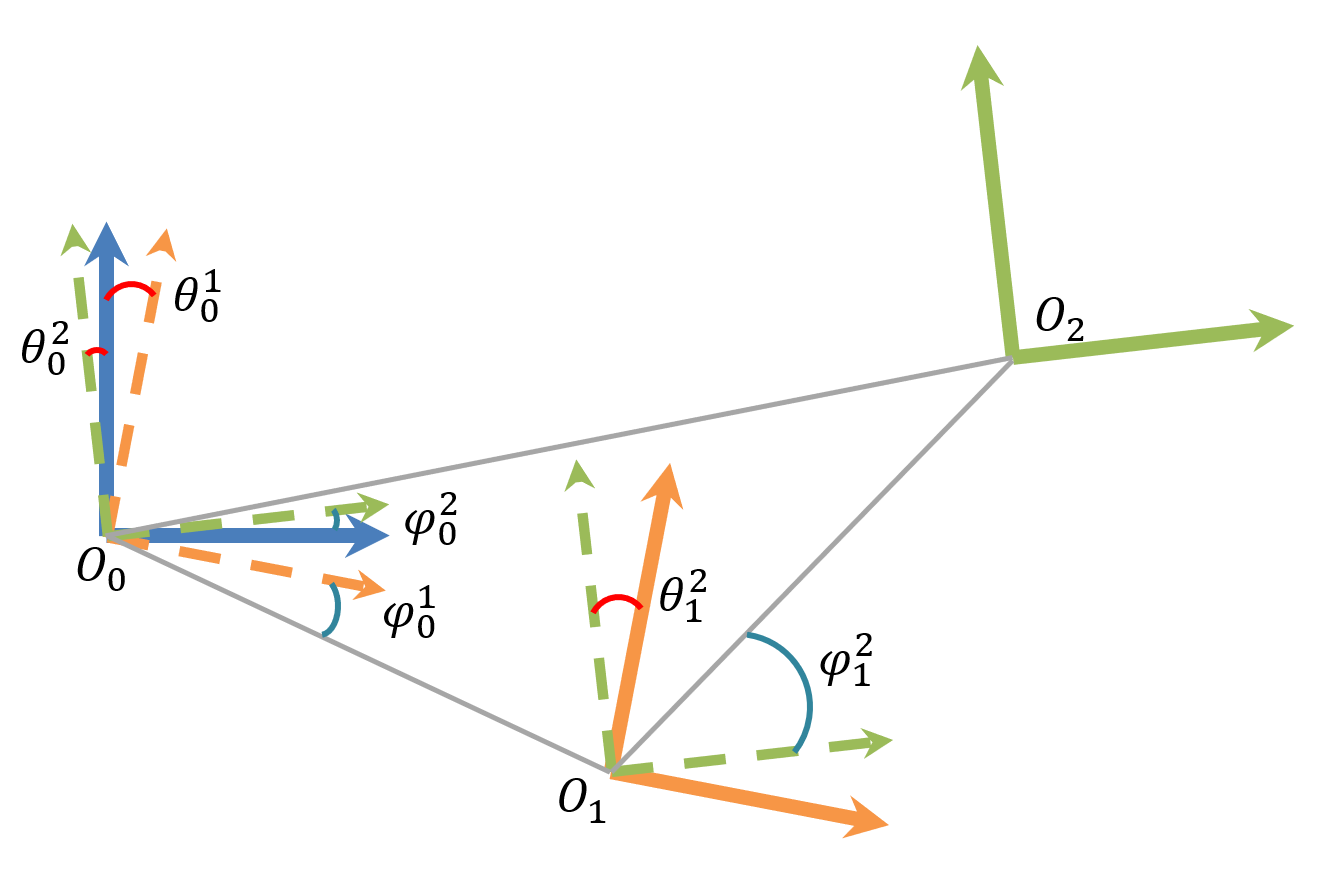}
    \caption{Translation constrain.}
    \label{fig:trans_constrain}
\end{figure}

\subsection{Optimization}\label{sec:Opti}
For this optimization, for three consecutive frames, after the front-end VO algorithm estimates the transforms between the first two frames and the last two frames, another transform between the first and third frame is considered, thus forming a constraint which allows adjusting the transform between two consecutive frames.
As a matter of fact, among all transformations, we can only determine the rotation $\theta$ and translation direction $\varphi$, but not the exact zoom and translation length between two frames. Instead, we can get the zoom energy vector and the translation energy vector. For every three consecutive frames $I_0, I_1$ and $I_2$, we do three image registrations: $I_0, I_1$, $I_1, I_2$ and $I_0, I_2$.
We set the transformation between the first and second frame as the \textbf{\textit{unit zoom}} $s_0^1$, \textbf{\textit{unit translation length}} $\rho_0^1$. Then we can denote all zooms and translation lengths between other frames by a scale factor. Additionally, we build a loop between three frames as shown in the upper part of Fig.~\ref{fig:pose}. The loop serves as a  transformation constraint that should satisfy Fig.~\ref{fig:trans_constrain}. To explain in detail,
take the translation for example, the translation between $I_0, I_1, I_2$ should satisfy equation \ref{eq:translation_constrain}:
\begin{equation}
    \centering
    \begin{aligned}
        &{\lambda_{\mathcal{T}}}_{01}^{02}  \rho_0^1 \cos(\theta_0^2 + \varphi_0^2) = \\
        &\rho_0^1 \cos(\theta_0^1 + \varphi_0^1) + {\lambda_{\mathcal{T}}}_{01}^{12}  \rho_0^1 \cos(\theta_0^1 + \theta_1^2 + \varphi_1^2) \\ \\
        &{\lambda_{\mathcal{T}}}_{01}^{02}  \rho_0^1 \sin(\theta_0^2 + \varphi_0^2) =  \\
        &\rho_0^1 \sin(\theta_0^1 + \varphi_0^1) + {\lambda_{\mathcal{T}}}_{01}^{12}  \rho_0^1 \sin(\theta_0^1 + \theta_1^2 + \varphi_1^2) \\ 
    \end{aligned}
    \label{eq:translation_constrain}
\end{equation}
Similarly, scale and rotation also satisfy corresponding relations. For a loop composed of these three frames, the optimization function is:
\begin{equation}
    \centering
    \begin{aligned}
        err_{\mu} &= \| \rho_0^1 \cos(\theta_0^1 + \varphi_0^1) + {\lambda_{\mathcal{T}}}_{01}^{12}  \rho_0^1 \cos(\theta_0^1 + \theta_1^2 + \varphi_1^2) \\
        & - {\lambda_{\mathcal{T}}}_{01}^{02}  \rho_0^1 \cos(\theta_0^2 + \varphi_0^2) \| \\
        err_{\nu} &= \| \rho_0^1 \sin(\theta_0^1 + \varphi_0^1) + {\lambda_{\mathcal{T}}}_{01}^{12}  \rho_0^1 \sin(\theta_0^1 + \theta_1^2 + \varphi_1^2) \\
        & - {\lambda_{\mathcal{T}}}_{01}^{02}  \rho_0^1 \sin(\theta_0^2 + \varphi_0^2) \| \\
        err_{s} &= \| s_0^1 \cdot \frac{s_0^1}{{\lambda_{\mathcal{S}}}_{01}^{12}} \cdot \frac{{\lambda_{\mathcal{S}}}_{01}^{02}}{s_0^1} - 1 \| \\ 
    \end{aligned}
\end{equation}
The above local optimization is performed each step starting from the second frame.

\begin{figure*}[t]
\centering    
    \subfigure[Real dataset trajectories] {
         \label{fig:real_dataset}     
        \includegraphics[width=0.63\columnwidth]{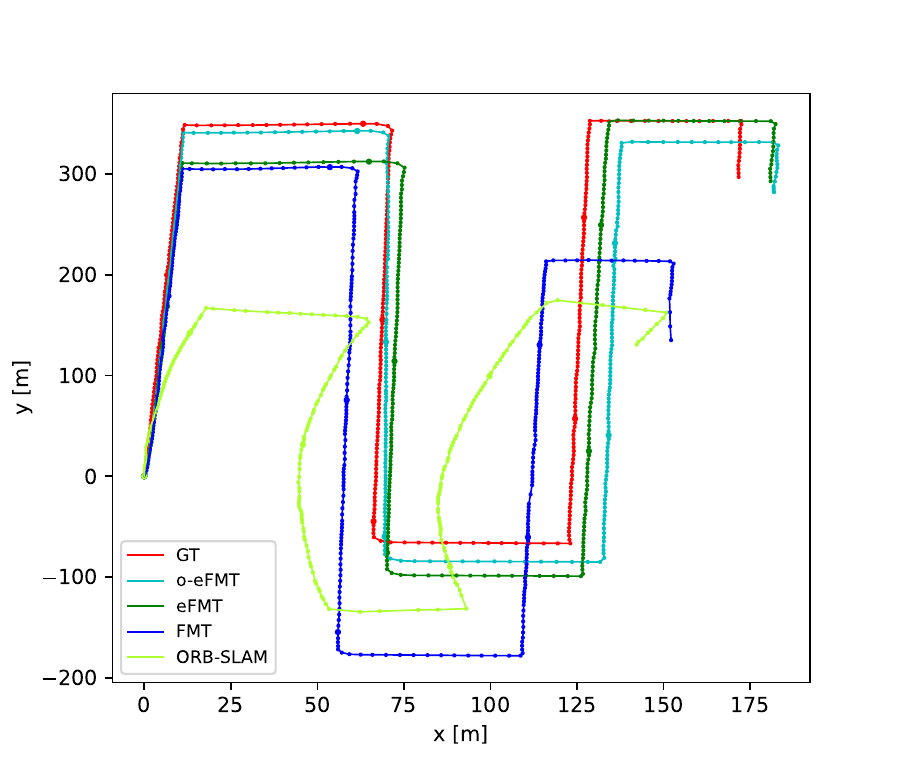}  
    } 
    \subfigure[Circle simulated dataset trajectories] {
         \label{fig:circle_dataset}     
        \includegraphics[width=0.63\columnwidth]{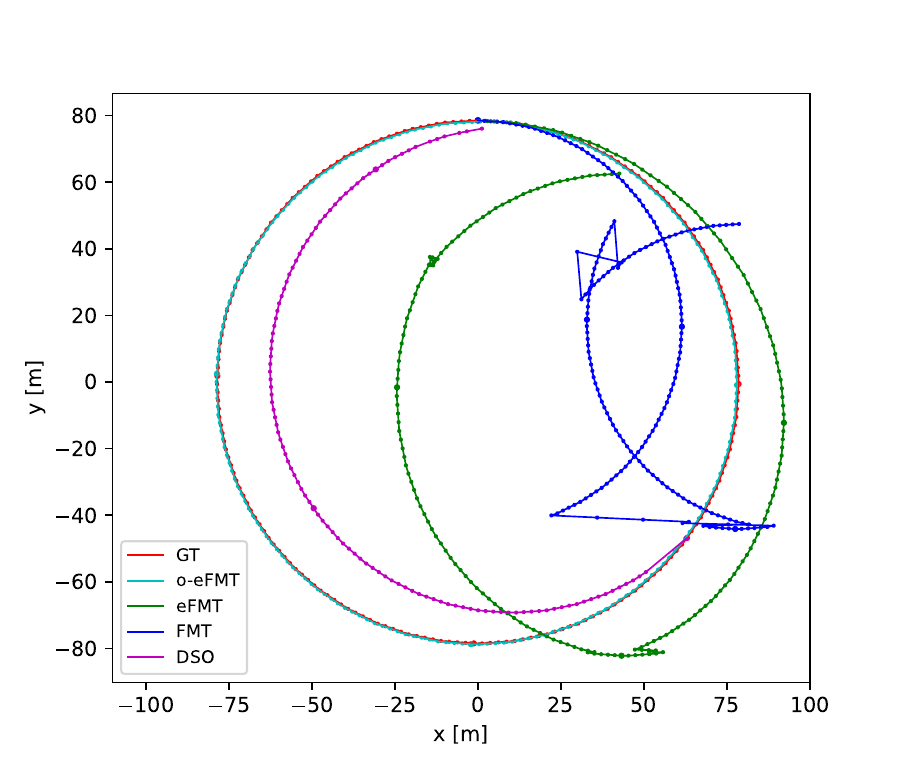}  
    }     
    \subfigure[Analemma simulated dataset trajectories] { 
        \label{fig:analemma_dataset}     
        \includegraphics[width=0.63\columnwidth]{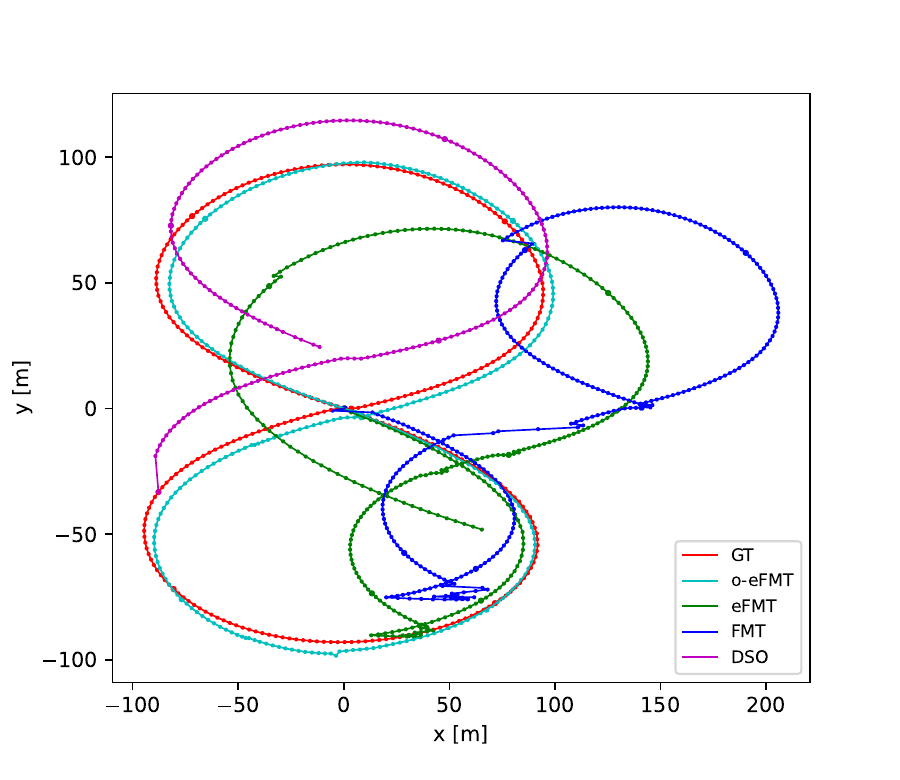}     
    }    
    \caption{Overall trajectories of different methods}     
    \label{fig:err_fig}     
\end{figure*}
\section{Experiments}
Our experiments are conducted on the same computer with i7-7700 CPU @ 3.60GHz × 8. We timed each registration for both o-eFMT and eFMT, which takes 0.1432s and 0.4069s on average, respectively. Our algorithm is significantly faster than eFMT. In the rest of this section, we present 
our experiments and results.

\subsection{Experiments on the Real Dataset}
The real dataset is the large real-world dataset ShanghaiTech Campus\footnote{
\url{https://robotics.shanghaitech.edu.cn/static/datasets/eFMT/ShanghaiTech_Campus.zip}}\cite{xu2021rethinking}. This dataset was collected with an Unmanned Aerial Vehicle (UAV) equipped with a down-looking camera and a DJI Matrice-300 RTK. The real scenario contains multi-depth planes such as building rooftops, ground, bridges and some other objects. The image capture frequency is 0.5 Hz and the RTK provides the groundtruth of the camera pose. Since it is a high-resolution dataset, we crop and resize it to 512 $\times$ 512 before experiments. 

For the experiments, we assume that the distances between adjacent poses are similar. For the same pose number results (o-eFMT, eFMT, FMT, ORB-SLAM3 and groundtruth), we align the initial pose of all the trajectories. Afterwards, upon obtaining the geometric lengths of different trajectories, we scale them based on the ratio between the trajectory length and the groundtruth trajectory length. If the pose number is smaller than the groundtruth (as in DSO), the scale is still based on the radio mentioned before, but afterwards, it is scaled again by multiplying the radio between the number of its poses and groundtruth. Then we translate the entire trajectory to minimize the error. 

After optimal scaling and registration, the absolute error of one trajectory is calculated by 
\begin{equation}
    \begin{aligned}
        err = \frac{1}{n} \sum_1^{n}{\|p_i - gt_i\|} \\
    \end{aligned}
\end{equation}
where n is the number of frames in this trajectory, \textit{p$_i$} is one trajectory pose and \textit{gt$_i$} is its corresponding groundtruth pose. The overall trajectories and the absolute trajectory errors are shown on Fig. \ref{fig:err_fig} and Table. \ref{tab:realError_tab}. 

Since DSO fails to track the camera pose, we do not show its trajectory in the figure. It shows that ORB-SLAM3 can estimate the pose in the beginning but with the error accumulation, the trajectory gradually deviates from the groundtruth. With the dominant depth plane changing, FMT will lose the real camera pose translation since the scale consistency is not accurate. We can see that our o-eFMT performs better than eFMT on this real dataset.

\subsection{Experiments on the Simulated Datasets}

\begin{table}[t]
    \centering
    \caption{Absolute trajectory error comparison on the real aerial dataset.}
    \label{tab:realError_tab}
    \begin{tabular}{lcccr}
    \toprule
        \textbf{VO methods} & \textbf{Max(m)} & \textbf{Mean(m)}& \textbf{Median(m)}\\ \midrule
        o-eFMT & \textbf{27.3} & \textbf{15.7} & \textbf{18.8} \\
        eFMT& 41.3 & 27.5 & 32.5  \\ 
        FMT & 163.2 & 79.8 & 86.3 \\
        ORB-SLAM3 & 339.8 & 165.3 & 159.7 \\
        DSO & $\backslash$ & $\backslash$ & $\backslash$ \\
    \bottomrule
    \end{tabular}
\end{table}

\begin{table}[t]
    \centering
    \caption{Absolute trajectory error comparison on the simulated datasets.}
    \label{tab:simError_tab}
    \begin{tabular}{lccccr}
    \toprule
        & \textbf{VO methods} & \textbf{Max(m)} & \textbf{Mean(m)}& \textbf{Median(m)}\\ \midrule
       & o-eFMT & \textbf{1.01} & \textbf{0.58} & \textbf{0.58} \\
      & eFMT & 55.7 & 34.2 & 44.3  \\ 
       Circle & FMT & 112.9 & 64.3 & 85.4 \\
       & ORB-SLAM3 & $\backslash$ & $\backslash$ & $\backslash$ \\
       & DSO & 42.2 & 25.6 & 26.7 \\
    \midrule
       & o-eFMT & \textbf{8.7} & \textbf{5.2} & \textbf{5.4} \\
     & eFMT& 98.0 & 51.4 & 54.6  \\ 
     Analemma    & FMT & 161.2 & 98.7 & 113.6 \\
       & ORB-SLAM3 & $\backslash$ & $\backslash$ & $\backslash$ \\
       & DSO & 41.4 & 27.8 & 25.3 \\
    \bottomrule
    \end{tabular}
\end{table}

\begin{figure}[t]
    \centering
   \includegraphics[width=1\linewidth]{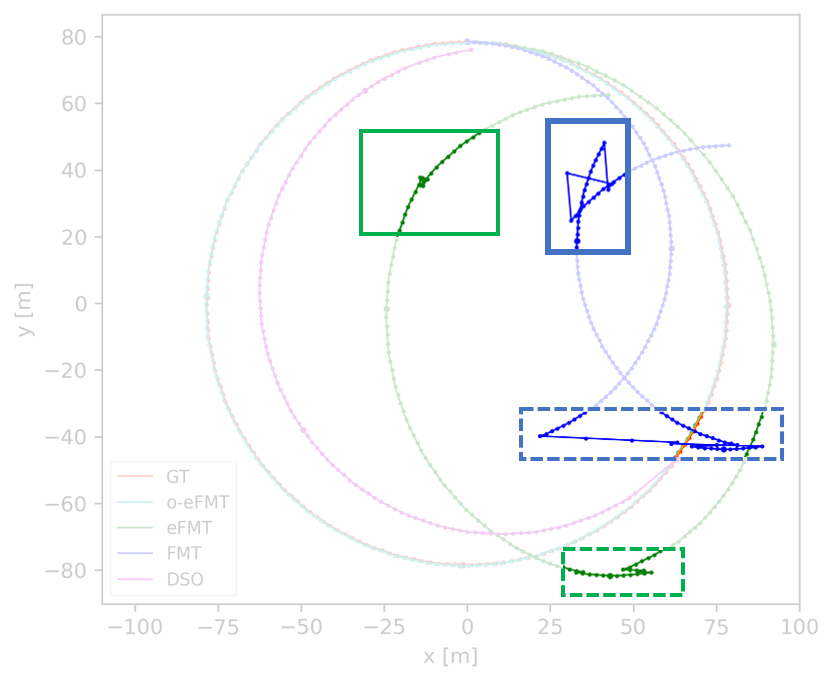}
    \caption{The blocks show the non-consecutive parts of FMT and eFMT trajectories. The solid frames and dotted frames correspond to two depth-change situations, respectively.}
    \label{fig:fragment}
\end{figure}

In this case, we use Blender to generate a simulation scenario, shown in Fig. \ref{fig:scenario}. This scenario mainly has three different depth planes, each of which are made up of multiple blocks. Based on their distance from the camera, we call them background, far plane and near plane. These three planes have different textures. Background and far plane are parallel to the $XY$ plane and the whole near plane has a 6 degrees inclination. This scenario has different depth planes and inclined planes, so we can use it to present experiments, and compare our o-eFMT with FMT and some other VO methods. The camera and different shapes of camera tracks are also simulated. The tracks are all in the plane which is parallel to the background. Meanwhile, the camera will run along the specified track and record pictures as the dataset.

After setting the frame number, the camera will move at a constant speed and record until it arrives at the end point of the track. At the same time, during the recording process, the real pose of the camera in the world coordinate system will also be recorded and exported. As Fig. \ref{fig:scenario} shows, there is a circled camera track above the whole scenario. The camera runs along the track and records multiple pictures with the set resolution as the blue frame shows. The camera is a pinhole camera, which has fixed intrinsics. The whole lengths of the circle track is 491 meters and of the Analemma track is 880 meters. The circle track dataset is 200 frames and the analemma one is 300 frames.

Since the camera only moves in the $XY$ plane, we can ignore the position in the $z$-axis. Fig. \ref{fig:err_fig} compares the localization results with different methods, including groundtruth (red), o-eFMT (cyan), eFMT (green), FMT (blue) and DSO (purple). ORB-SLAM3 fails to estimate the camera pose since the features in this scenario are highly similar, which means that the feature descriptor cannot distinguish the feature points and the feature matching will fail. The absolute error is shown in Table \ref{tab:simError_tab}.

It can be concluded that DSO always fails to generate a trajectory in the beginning. This is because DSO depends too much on the first few images. If the images show different planes, DSO always chooses to reset the initial map until it gets some consecutive frames that have similar planar textures. In the circle track, o-eFMT, eFMT and FMT all represent the overall trajectories. o-eFMT is very close to the groundtruth. We mainly focus on the eFMT and FMT trajectories fragments shown in Fig. \ref{fig:fragment}. We discover that in the blocks that we marked out, the trajectories of FMT and eFMT are non-consecutive, especially FMT. This is because the depths of the corresponding images change. Scale consistency cannot be guaranteed since FMT always finds the maximum value in the PSD. As Fig. \ref{fig:PatternMatching} shows, the peak value changes a lot, however, the camera pose translation does not change too much. Compared with FMT and eFMT, o-eFMT improves the accuracy in this multi-depth scenario significantly. We believe that eFMT is affected by noise during energy vector extraction and pattern matching.

The analemma track is a more challenging and complex scenario. There is less overlap between two adjacent frames, which is the main reason for all VO methods being less robust than before. We can see that o-eFMT makes some errors, but its trajectory is still similar to the groundtruth. eFMT has larger errors since encountering the first mis-registration. All other methods have failed to track the the camera pose.


\section{Conclusions}
This paper proposed the optimized eFMT algorithm. We introduced new ways to extract both scale and rotation energy vectors from the PSD, and used improved pattern matching on the energy vectors to determine transformations amongst three consecutive frames. To improve the accuracy we added a back-end optimization to our version of eFMT, which adds local constraints, which not only improves accuracy but also boosts the robustness of the entire framework. Our approach is significantly faster than eFMT, at the cost of some accuracy. But this is offset by the other improvements to the algorithm presented here. Our experiments show the superior accuracy of o-eFMT over eFMT, FMT and the traditional methods ORB-SLAM3 and DSO.

For future work we aim to further improve the algorithm by adding loop-closure detection to provide more efficient data for the back-end optimization and thus turn it into a real SLAM algorithm. This will lead to a global consistent camera pose estimation and further robustness improvements. In the upcoming journal paper we will also do a more thorough investigation of the depth-filtering effect of the scale, which we think can lead to an even improved algorithmic approach. We also plan to integrate our algorithm into the SLAM Hive benchmarking suite \cite{yang2023slamhive} for a more thorough evaluation. 

\section*{ACKNOWLEDGMENTS}

This work has been partially funded by the Shanghai Frontiers Science Center of Human-centered Artificial Intelligence and it was supported by Science and Technology Commission of Shanghai Municipality (STCSM), project 22JC1410700 "Evaluation of real-time localization and mapping algorithms for intelligent robots".










\IEEEtriggeratref{15}

\bibliographystyle{IEEEtran}
\bibliography{IEEEabrv,Bibliography}

\end{document}